\documentclass[sigconf]{acmart}
\AtBeginDocument{%
  }

\setcopyright{acmlicensed}
\copyrightyear{2018}
\acmYear{2018}
\acmDOI{XXXXXXX.XXXXXXX}
\acmConference[Conference acronym 'XX]{Make sure to enter the correct
  conference title from your rights confirmation email}{June 03--05,
  2018}{Woodstock, NY}
\acmBooktitle{Companion Proceedings of the 34th ACM Symposium on the Foundations of Software Engineering (FSE ’26), July 5 - 9, 2026 Montreal, Canada}

\acmISBN{978-1-4503-XXXX-X/2018/06}

\usepackage{enumitem}
\usepackage{latexsym}
\usepackage{algorithm}
\usepackage[noend]{algorithmic}
\usepackage{fontawesome}
\usepackage{pifont}
\usepackage{listings}[language=Python]
\usepackage{graphicx}
\usepackage{booktabs}
\usepackage{mdframed}
\usepackage{lipsum}
\usepackage{placeins}
\usepackage{multirow}
\usepackage{amsmath}
\usepackage{hyperref}
\usepackage{cancel}
\usepackage{array}
\usepackage[subtle]{savetrees}
\usepackage{xspace}
\usepackage{xcolor}

\usepackage{amssymb}

\definecolor{mygray}{gray}{0.97}
\colorlet{shadecolor}{mygray}

\newmdenv[%
  backgroundcolor=mygray, 
  linewidth=0pt
]{newshaded}

\definecolor{codegreen}{rgb}{0,0.6,0}
\definecolor{codegray}{rgb}{0.5,0.5,0.5}
\definecolor{codepurple}{rgb}{0.58,0,0.82}
\definecolor{backcolour}{rgb}{0.95,0.95,0.92}

\definecolor{mauve}{rgb}{0.58,0,0.82}

\definecolor{ForestGreen}{HTML}{029147}

\lstdefinestyle{mystyle}{
    language=Python,
    backgroundcolor=\color{backcolour},   
    commentstyle=\color{codegreen},
    keywordstyle=\color{blue},
    numberstyle=\tiny\color{codegray},
    stringstyle=\color{codepurple},
    basicstyle=\ttfamily\footnotesize,
    breakatwhitespace=false,         
    breaklines=true,                 
    captionpos=b,                    
    keepspaces=true,                 
    numbers=left,                    
    numbersep=5pt,                  
    showspaces=false,                
    showstringspaces=false,
    showtabs=false,                  
    tabsize=2,
    columns=fullflexible,
}

\DeclareMathOperator*{\argmax}{arg\,max}

\begin{document}

\title{Reasoning Trajectories for Socratic Debugging of Student Code:\\From Misconceptions to Contradictions and Updated Beliefs}


\author{Erfan Al-Hossami}
\email{ealhossa@charlotte.edu}
\affiliation{%
    \institution{University of North Carolina at Charlotte}
    \country{Charlotte, NC, USA}
}

\author{Razvan Bunescu}
\email{rbunescu@charlotte.edu}
\affiliation{%
  \institution{University of North Carolina at Charlotte}
   \country{Charlotte, NC, USA}
}






\renewcommand{\shortauthors}{Al-Hossami et al.}

\begin{abstract}
In Socratic debugging, instructors guide students towards identifying and fixing a bug on their own, instead of providing the bug fix directly. Most novice programmer bugs are caused by programming misconceptions, namely false beliefs about a programming concept. In this context, Socratic debugging can be formulated as a guided Reasoning Trajectory (RT) leading to a statement about the program behavior that contradicts the bug-causing misconception. Upon reaching this contradiction, the ensuing cognitive dissonance is expected to lead the student to identify the false belief on their own, followed by an enduring belief update. In this paper, we introduce the task of reasoning trajectory generation, together with a dataset of debugging problems annotated with RTs that are manually created or LLM-generated. We then describe LLM-based solutions for generating RTs and Socratic conversations that are anchored on them. A large scale LLM-as-judge evaluation shows that large language and reasoning models can generate up to 91\% correct reasoning trajectories and 98.7\% valid conversation turns.
\end{abstract}



\received{20 February 2007}
\received[revised]{12 March 2009}
\received[accepted]{5 June 2009}

\maketitle

\section{Introduction and Motivation}
\label{sec:introduction}

One of the most effective ways of improving students' learning is through {\it Socratic dialogue}. In a Socratic dialogue, a teacher guides a learner within their zone of proximal development~\cite{vygotsky2012thought} by asking questions and providing feedback, with the purpose of directing them towards solving a problem on their own rather than providing solutions directly. The instructor's questions may probe a student’s existing knowledge or assumptions; guide attention to relevant aspects of a complex problem; or encourage discovery of general principles through the consideration of alternative solutions or counterexamples~\cite{elder_role_1998}. Through its emphasis on active inquiry, in-context reasoning about evidence, and repeated retrieval of relevant concepts from memory, Socratic dialogue engages students in deep thinking and meaningful integration of new knowledge, which can greatly improve their acquisition of generalizable skills and ultimately their learning outcomes \cite{brown:cacm24}. 

Socratic questioning is often used in instructional scaffolding~\cite{quintana2018scaffolding} and is effective in enhancing learning gains in code comprehension tasks~\cite{tamang2021_socratic}. While Socratic questioning can substantially improve learning outcomes, it is time-consuming and cognitively demanding, requiring human instructors to continuously assess a student's understanding and to tailor questions to be most effective at each turn. 
In this paper, we introduce tools that support instructors to first {\it plan} and then {\it articulate} Socratic conversations, in the context of helping students fix buggy code. A significant part of the curriculum in beginner programming classes is allocated to programming exercises, where students are asked to solve coding problems with increasing levels of difficulty. However, when students learn to code they often develop false beliefs about various programming concepts, i.e., misconceptions, which can lead to buggy code. We assume an ideal scenario where the Student has access to an Instructor, such that when the student cannot fix the bug on his own, he seeks help from her\footnote{The genders for the Student and the Instructor roles were assigned by coin flip.}. The Instructor is assumed to be a proficient programmer, with experience in teaching novice programmers. When contacted by the Student for help, she aims to optimize learning by following a Socratic approach, where over one or more dialogue turns, she guides the student towards figuring out on his own the misconception causing the bug. Henceforth, we use the term {\sc Socratic debugging} \cite{erfan2023_bea,erfan_2024_sigcse} to refer to the ensuing dyadic conversation between a Student and an Instructor.

\begin{figure*}[t]
\centering
    \includegraphics[width=1.0\textwidth]{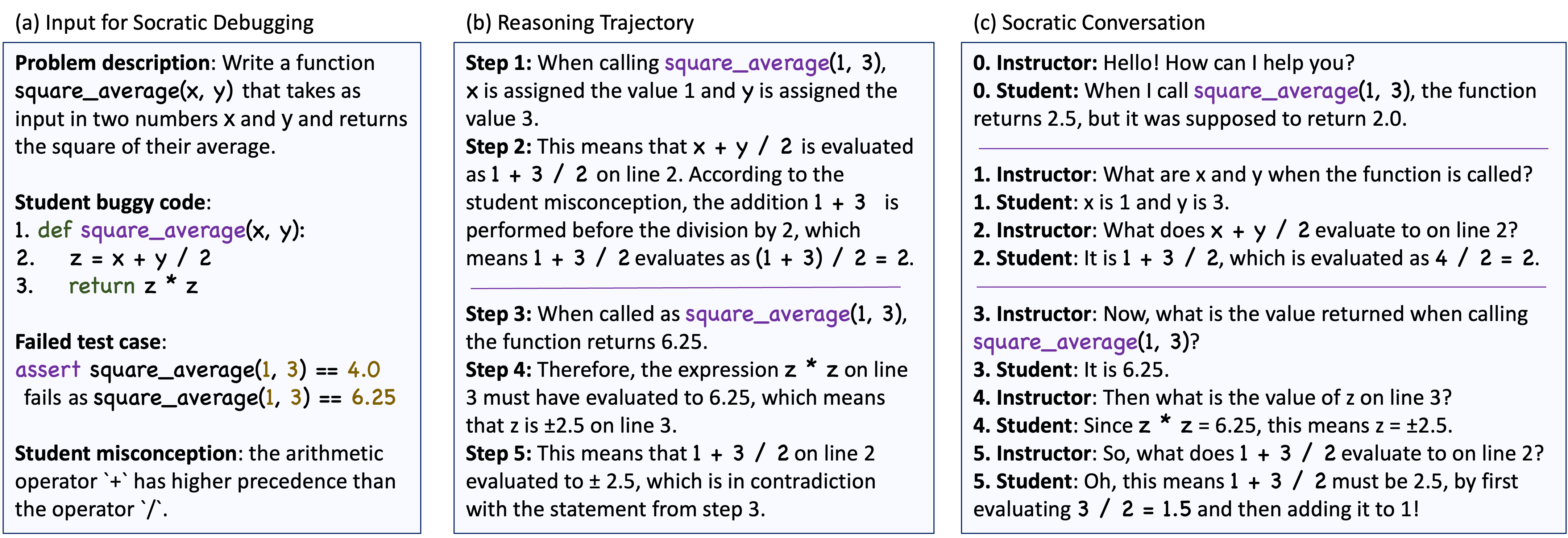}
    \caption{Socratic debugging example: (a) the input specifies the problem, the buggy code, the failed test case, and the student misconception that caused the bug; (b) a reasoning trajectory ending with a statement that contradicts the misconception; (c) a Socratic conversation that follows the reasoning trajectory and ends with a belief update.}
    \label{fig:task}
\end{figure*}

In a Socratic debugging approach, it is essential that the student himself realizes which of his programming beliefs are false, i.e., misconceptions. By guiding the student to discover and fix a misconception on his own, the instructor also maximizes the likelihood that the fixed belief will endure over time and not revert to the initial false belief. To achieve this aim, we propose that the Instructor guide the student along a sequence of inferences about the code behavior for a failed test case. The {\it reasoning trajectory} is designed such that the last inference step proves a statement that is in direct contradiction with the student's misconception. This overt contradiction between the false belief and the actual code behavior is expected to create a strong {\it cognitive dissonance} \cite{festinger1959cognitive} for the student, who consequently not only realizes which of his beliefs is false, but also corrects it on his own, as shown in the example in Figure~\ref{fig:task}. In general, the psychological discomfort associated with cognitive dissonance has been found empirically to be extremely motivating in terms of triggering learning processes that seek to resolve the dissonance \cite{zanna1976dissonance,elliot1994motivational}. As described in \cite{adcock_2012_dissonance_learning}, placing learners in a state of cognitive dissonance is ideal for learning in problem-solving scenarios, where the intrinsic human need for consistency and equilibrium leads to a constant process of examining new information and updating existing knowledge structures \cite{piaget1985equilibration}.

The rest of the paper is structured as follows: in Section~\ref{sec:task} we provide a definition of reasoning trajectories and Socratic conversations that are anchored in them; in Section~\ref{sec:dataset} we describe the benchmark dataset created to support the development and evaluation of LLM-based approaches for articulating RTs and Socratic turns, which we introduce in Sections~\ref{sec:rt} and~\ref{sec:sc}, respectively; in Section~\ref{sec:experiments} we present and discuss experimental results, whereas in Section~\ref{sec:related} we summarize related work. The paper ends with conclusion and thoughts on future work in Section~\ref{sec:conclusion}.

\section{Task Definition}
\label{sec:task}



\begin{figure}[t]
\centering
    \includegraphics[width=1.0\columnwidth]{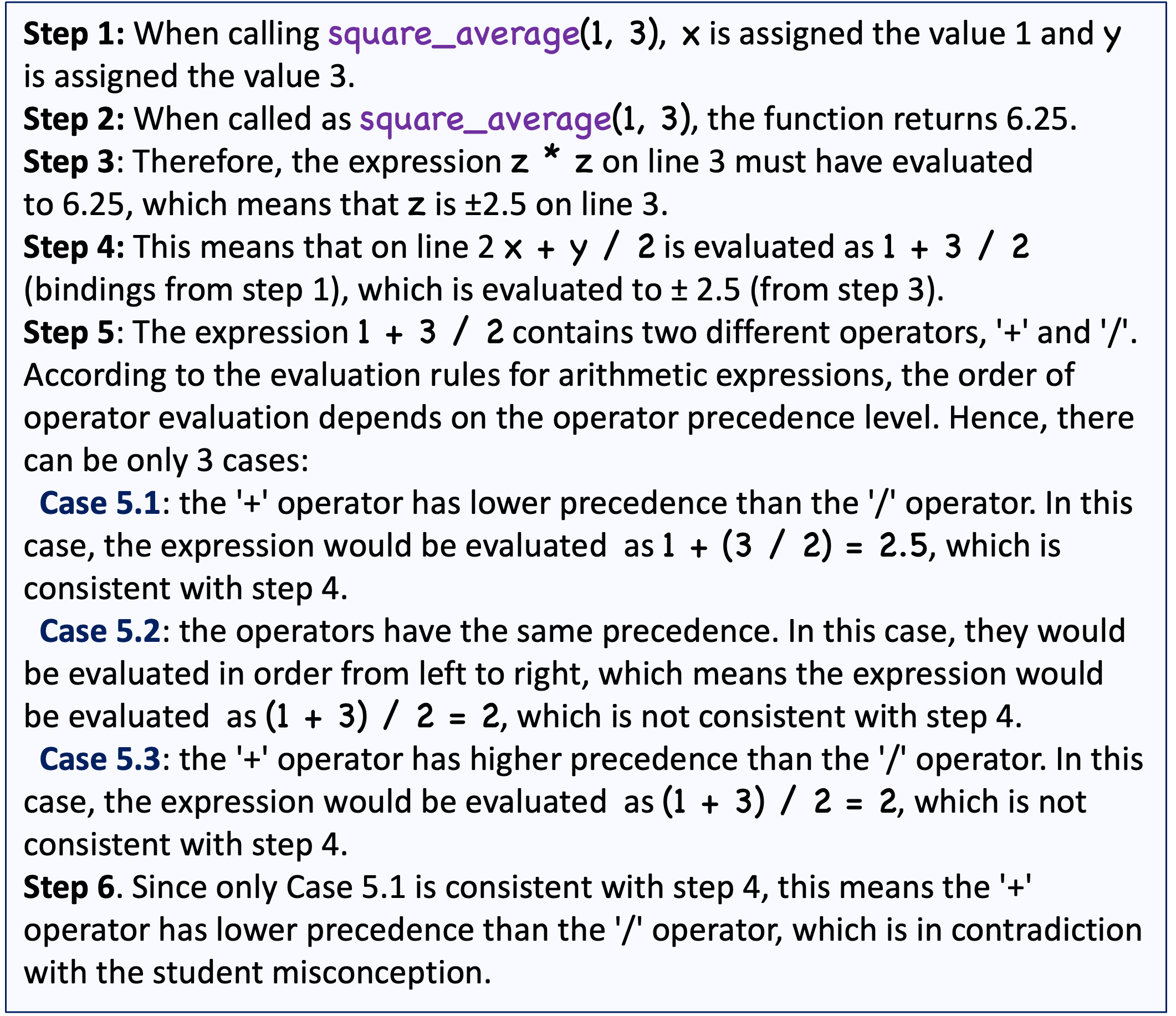}
    \caption{Alternative reasoning trajectory for the input from Figure~\ref{fig:task}(a).}
    \label{fig:rt-alt}
\end{figure}

As shown in Figure~\ref{fig:task}(a), the input to the Socratic debugging task consists of a problem description, the student's buggy code, a failed test case, and the student's misconception that caused the bug. Consistent with the aforementioned aim of guiding the student towards discovering his own misconception, we approach the task of Socratic debugging as a pipeline of two main subtasks:
\begin{enumerate}[leftmargin=1em]
    \item {\bf Reasoning Trajectory (RT)}: In the first step, a reasoning trajectory is generated as a sequence of inference steps such that the statement proven in the last step contradicts the misconception or provides a counterexample to the misconception, as shown in Figure~\ref{fig:task}(b).
    \item {\bf Socratic Conversation (SC)}: In the second step, a Socratic conversation is generated step by step, such that each RT step is associated with an Instructor turn followed by a Student turn, where the instructor's question aims to elicit from the student the statement proven at that step, as illustrated in Figure~\ref{fig:task}(c).
\end{enumerate}
The reasoning trajectory shown in Figure~\ref{fig:task}(b) is structured in two parts. In the first part, the reasoning steps lead to showing a statement of the student's misconception for the failed test case, namely that the expressions 1 + 3 / 2 evaluates as 2. In the second part, the reasoning proceeds backwards from the returned value in order to infer a statements that contradicts the misconception statement, namely that 1 + 3 / 2 evaluates to $\pm 2.5$. Note that this is not the only way of articulating an RT that ends with a statement contradicting the misconception. Figure~\ref{fig:rt-alt} shows an alternative RT where the reasoning steps end with a statement that is the opposite of the misconception statement.
Thus, while the RT from Figure~\ref{fig:task} (b) can be seen as providing a counterexample to the misconception statement by instantiating it for a particular failed test case, the RT in Figure~\ref{fig:rt-alt} does not instantiate the misconception statement and instead proves a statement that contradicts the misconception statement in the general case. Given that the first type of RTs are generally shorter, in this paper we focus on generating RTs that derive counterexamples to the student misconception. The LLM-based approach for generating reasoning trajectories is described in Section~\ref{sec:rt}.

Once a reasoning trajectory is generated, it is used step by step to generate a corresponding Socratic conversation. As shown in Figure~\ref{fig:task}(c), the Socratic conversation is structured in three parts. The first part contains a generic, initial statement from the instructor, while the student's turn describes the failed test case. The turns in the second and third parts map to the steps in the first and second parts of the RT, respectively. For each RT step, the instructor asks a question that aims to guide the student towards articulating the statement proven at that step. Note that although we generate a Socratic turn for each RT step, it is possible for the instructor to skip one or more steps if she determines that the student is capable of making one or more inferences on his own, without guidance. For example, right after turn 3, the instructor can choose to skip turn 4 and go directly to turn 5. The LLM-based approach for generating reasoning trajectories is described in Section~\ref{sec:sc}.


\begin{figure}[t]
\centering
    \includegraphics[width=1.0\columnwidth]{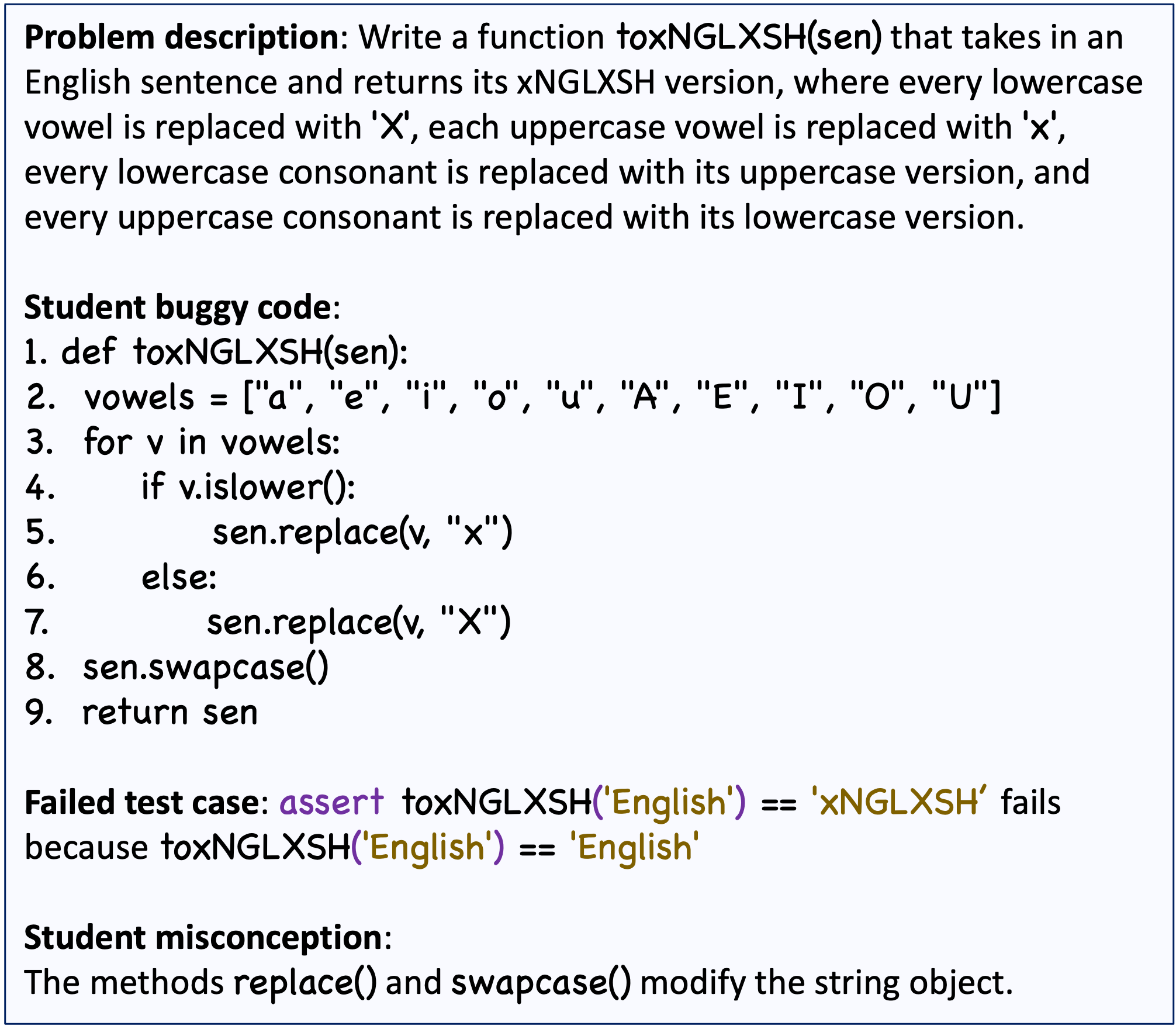}
    \caption{The original input specification.}
    \label{fig:original}
\end{figure}

\subsection{Simplification}

\begin{figure}[t]
\centering
    \includegraphics[width=1.0\columnwidth]{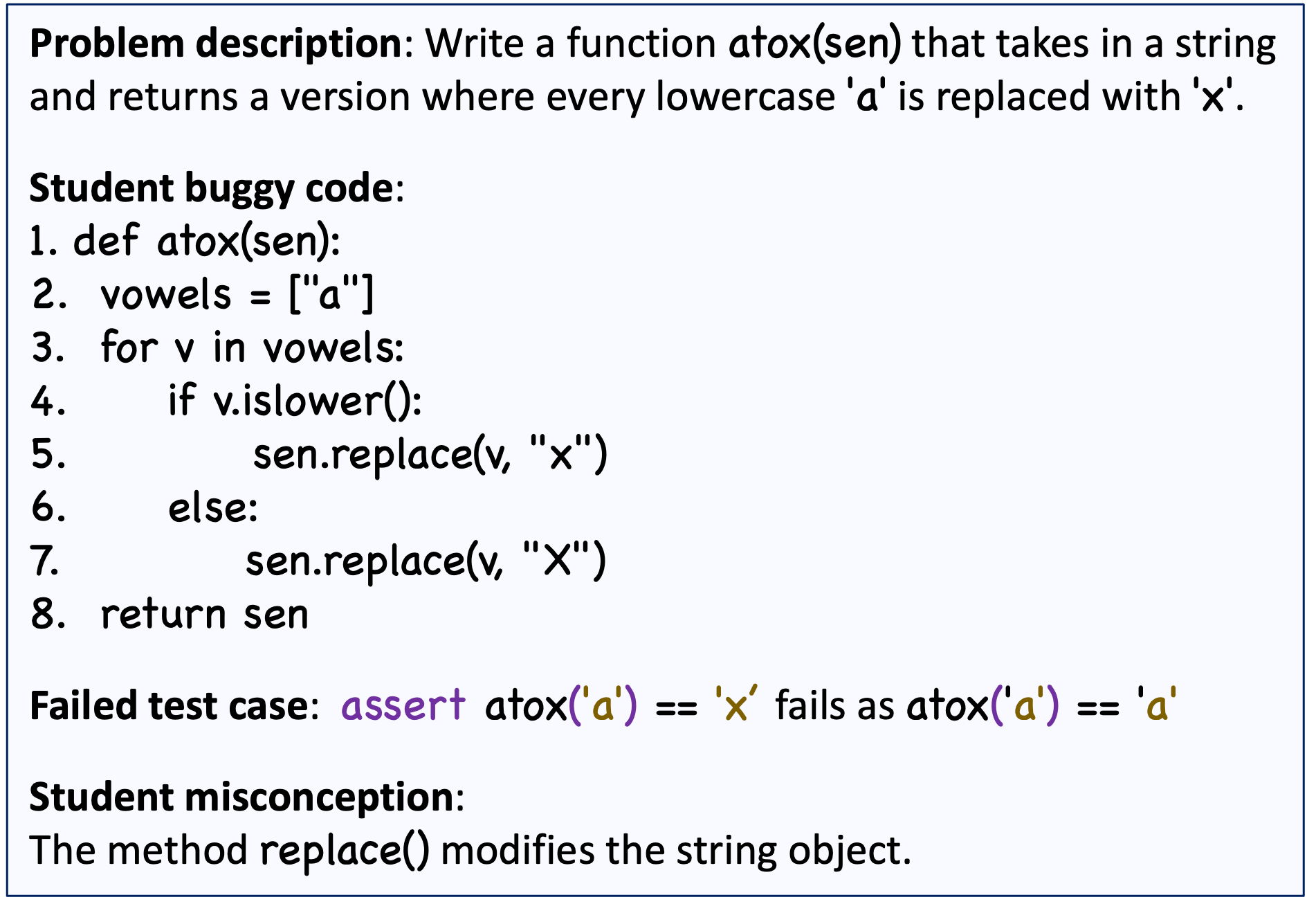}
    \caption{The simplified input for the original in Figure~\ref{fig:original}.}
    \label{fig:simplification}
\end{figure}

\begin{figure}[!t]
\centering
    \includegraphics[width=1.0\columnwidth]{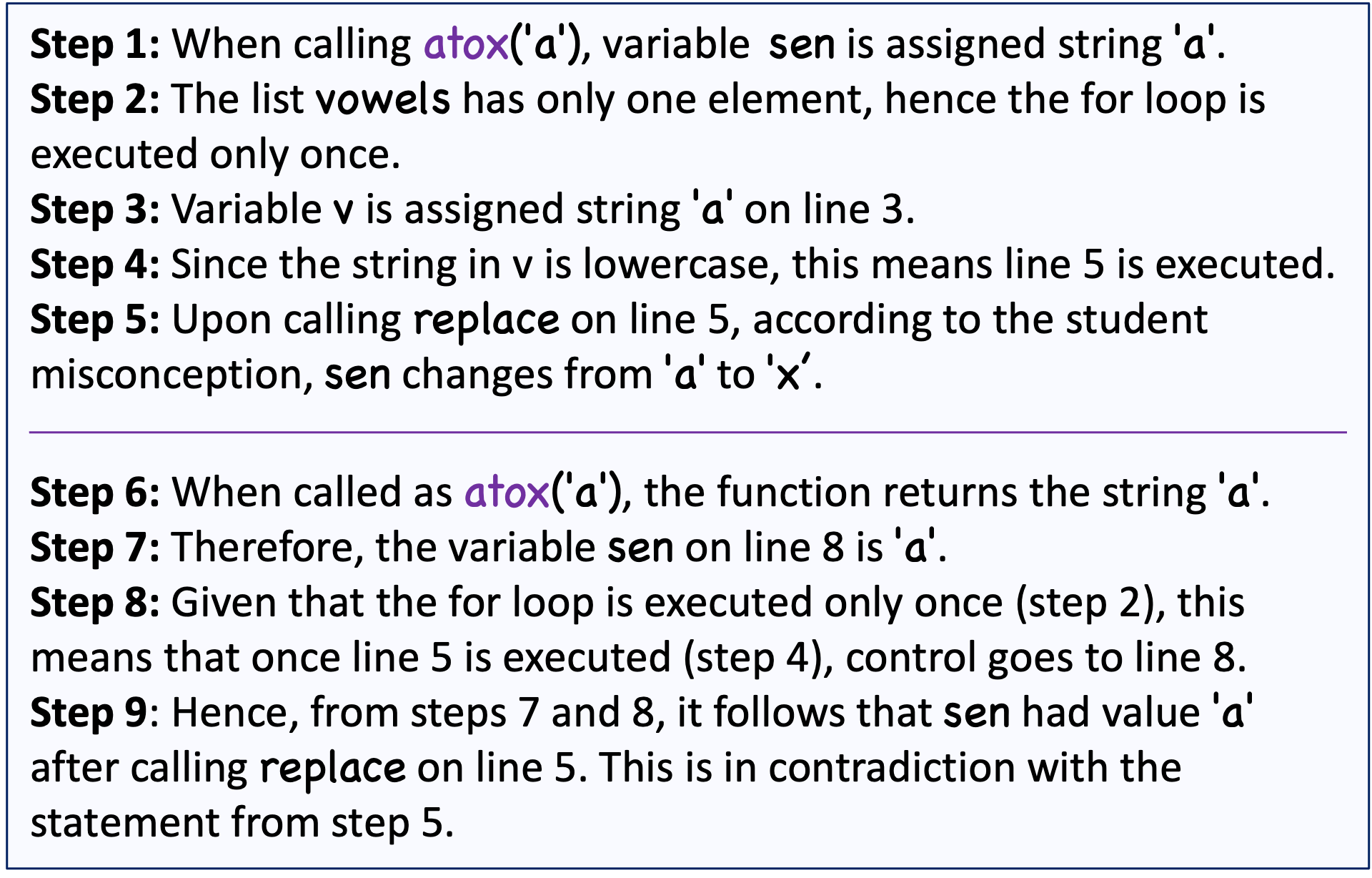}
    \caption{The RT for the simplified input in Figure~\ref{fig:simplification}.}
    \label{fig:rt-simple}
\end{figure}

As shown in Figure~\ref{fig:task}(b), the reasoning trajectory is structured in two parts: the first part leads to an instance of the misconception, whereas the second part leads to a statement that contradicts it. It is important for both parts in this reasoning process to be short, otherwise a long and complicated reasoning trajectory can place a significant cognitive burden on the student, which will defeat the aim of Socratic guidance. Therefore, to keep the complexity of the reasoning traces at a feasible level, we envision a simplification process where the original problem description, code, and failed test case are simplified such that (a) they focus on the code behavior that instantiates the misconception, while (b) they stay as close to the original as possible. In Figure~\ref{fig:original} we show an example input, where formulating a reasoning trajectory would be overly complicated due to the many calls to the function about which the student has a misconception, and the length of the input string. Furthermore, the student has two misconceptions, whereas by definition a reasoning trace corresponds to just one misconception. While it is possible to merge the two misconceptions into a general misconception that subsumes both, e.g., "string methods can modify the string object", it is easier for the student to address concrete misconceptions, one at a time. Correspondingly, the original task is simplified as shown in Figure~\ref{fig:simplification}, whereas the corresponding RT is shown in in Figure~\ref{fig:rt-simple}.


\section{Dataset}
\label{sec:dataset}

We leverage a subset of the dataset of problems, their coding solutions, and misconceptions that was released with the {\sc McMining} benchmark described in~\cite{erfan_2025_mcmining},  containing 501 problems, 558 coding solutions, and 40 bug-inducing misconceptions.\footnote{All code and data will be released publicly upon acceptance.} We then use an LLM to corrupt solution code such that it exhibits a misconception. To ensure naturalness and plausibility in LLM-generated corrupted code, we pair each misconception with solutions that contain the necessary programming constructs. This is achieved through the construct-based pairing Algorithm~\ref{alg:pairing}, which takes as input the set of coding solutions and misconceptions, and outputs the most suitable solutions that rely on the programming concept referenced by the misconception. 
\begin{algorithm}[t]
\caption{Construct-Based Misconception-Solution Pairing}
\label{alg:pairing}
\begin{algorithmic}[1]
\REQUIRE Solutions $S$ with extracted constructs,\\ Misconceptions $M$ with constructs, Target count $N$
\ENSURE Pairings $\mathcal{D} = \{(m_i, s_i)\}_{i=1}^{N}$

\STATE \textbf{Phase 1: Extract Constructs}
\FOR{each solution $s \in S$}
    \STATE $\text{Constructs}[s] \gets \text{ExtractConstructs}(s)$ \COMMENT{AST + regex}
\ENDFOR

\STATE \textbf{Phase 2: Generate Pairs}
\STATE $\text{Used} \gets \emptyset$ \COMMENT{track used solutions}
\STATE $\mathcal{D} \gets \emptyset$ \COMMENT{result pairings}
\STATE $i \gets 0$ \COMMENT{round-robin index}

\WHILE{$|\mathcal{D}| < N$}
    \STATE $m \gets M[i \bmod |M|]$ \COMMENT{current misconception}
    \STATE $\text{Candidates} \gets \emptyset$
    
    \FOR{each solution $s \in S \setminus \text{Used}$}
        \STATE $\text{score(s)} \gets \left|\text{Constructs}[m] \cap \text{Constructs}[s]\right|$
        \IF{$\text{score(s)} \geq 1$ \OR $\text{IsSpecialCase}(m, s)$}
            \STATE $\text{Candidates} \gets \text{Candidates} \cup \{\langle s, \text{score(s)}\rangle\}$
        \ENDIF
    \ENDFOR
    
    \IF{$\text{Candidates} \neq \emptyset$}
        \STATE $\displaystyle \hat{s} \gets \argmax_{\langle s, \text{score(s)}\rangle \in \text{Candidates}} \text{score(s)}$
        \STATE $\mathcal{D} \gets \mathcal{D} \cup \{(m, \hat{s})\}$
        \STATE $\text{Used} \gets \text{Used} \cup \{\hat{s}\}$
    \ENDIF
    
    \STATE $i \gets i + 1$
\ENDWHILE

\RETURN $\mathcal{D}$
\end{algorithmic}
\end{algorithm}
The pairing algorithm operates in two phases: 
\begin{enumerate}
    \item Extract programming constructs from each solution using AST parsing and pattern matching, identifying over 80 distinct construct types.
    \item Generate (misconception, solution) pairs through semantic alignment based on shared coding construct.
\end{enumerate}
For each misconception $m$ annotated with related constructs, we compute an overlap score with each solution $s$ as $\text{score}(m, s) = |\text{constructs}[m] \cap \text{constructs}[s]|$. We accept pairs with $\text{score}(m, s) \geq 1$ for regular misconceptions. Special handling was needed for misconceptions where construct overlap was not sufficient, such as recursion misconceptions that require actual recursive function calls, or operator precedence misconceptions that require specific operator combinations. Algorithm~\ref{alg:pairing} employs a round-robin allocation, cycling through misconceptions while selecting the highest-scoring unused solution for each. This ensures diversity, where each solution is used at most once, while handling varying construct availability. The process is fully deterministic and uses standard Python libraries, such as \texttt{ast}, \texttt{re}, \texttt{json}.

Each solution in the $\langle${\it solution}, {\it misconception}$\rangle$ pairs created by Algorithm~\ref{alg:pairing} is mapped to the corresponding problem, and the resulting $\langle${\it problem}, {\it solution}, {\it misconception}$\rangle$ triplets are then used as input to the {\sc McInject} tool from~\cite{erfan_2025_mcmining}, which generates buggy code samples by injecting the misconception in the correct solution. To ensure the misconception is correctly exhibited in the buggy code, we use {\sc McInject} with up to 3 refinement iterations. The refinement process uses an LLM-as-judge to determine whether the buggy code exhibits a misconception or not, providing feedback to {\sc McInject} if the code does not yet exhibit the misconception. When used in this way, {\sc McInject} generated 250 corrupted code samples. Of these, 43 samples were filtered out: 17 due to passing all the unit tests and 26 due to not exhibiting the misconception. To the remaining 207 samples we added 20 manually designed samples, yielding a final dataset of 227 buggy code samples exhibiting an associated misconception. For each buggy code sample, we use an LLM (Claude Sonnet-4.5 with temperature 0.1, disabled reasoning, and max\_tokens 4000) connected to a code execution tool to identify and describe the simplest test case that the buggy code fails.

The problem description, buggy code sample, failed test case description, and misconception description were then used as input for the Socratic debugging pipeline (Section~\ref{sec:pipeline}). A reasoning trajectory is generated first (Section~\ref{sec:rt}), which is then used as input for generating a Socratic conversation (Section~\ref{sec:sc}). We use 14 different LLM configurations to generate reasoning trajectories and Socratic conversations for all 227 triplets. The overall statistics of the dataset are summarized in Table~\ref{tab:benchmark-stats} below.

\begin{table}[h]
\centering
\begin{tabular}{lr}
\toprule
\textbf{Component} & \textbf{Count} \\
\midrule
Problems & 501 \\
Solutions & 558 \\
Misconceptions & 40 \\
$\langle$Problem, Solution, Misconception$\rangle$ triplets & 227 \\
\midrule
\textbf{Manually created} & \\
\quad Reasoning Trajectories & 10 \\
\quad Total RT steps & 57 \\
\midrule
\textbf{LLM-Generated} & \\
\quad LLM configurations & 14 \\
\quad Total RT steps & 22,506 \\
\bottomrule
\end{tabular}
\caption{The breakdown of RT steps per each of the 14 LLM configurations is shown later in in Table~\ref{tab:rt-socratic-results}.}
\label{tab:benchmark-stats}
\end{table}

\section{Socratic Debugging Pipeline}
\label{sec:pipeline}

The generation of Socratic debugging conversations is implemented as a pipeline of two steps. First, a reasoning trajectory is generated that starts from the failed test case and ends with a correct statement about the buggy code behavior that contradicts the student's misconception (Section~\ref{sec:rt}). Then, the RT is used as a plan for generating a Socratic conversation, where each reasoning step is associated a Socratic turn composed of an Instructor utterance followed by a Student utterance (Section~\ref{sec:sc}).

\subsection{Reasoning Trajectories}
\label{sec:rt}

Given a problem description, the buggy code, a failed test case, and a misconception, an LLM is instructed to generate a sequence of deductive reasoning steps that culminate in a statement contradicting the student's false belief.

Figure~\ref{fig:prompt-rt} shows the prompt template used for RT generation.\footnote{Complete versions can be seen in the GitHub repository.} We employ a 2-shot prompting approach, which includes two worked examples and structured input and output formats. The prompt emphasizes five core principles that guide the generation process: (1) strict deductive reasoning with no logical leaps or abductive inferences; (2) consistency with the student's misconception, avoiding the use of programming knowledge that would contradict their false belief, e.g., if the student believes that \verb|range(n)| starts at 1, the RT should not use the fact that \verb|range(n)| actually starts at 0; (3) exclusive focus on contradicting the misconception rather than providing fixes; (4) starting from observable facts in the failed test case; and (5) sequential reasoning steps with explicit citation of premises.

These principles ensure that generated RTs maintain logical rigor and focus on deducing a statement that contradicts the misconception. Each inference step must follow necessarily from previously established facts and correct knowledge of Python programming that does not contradict the student's misconception. By requiring consistency with the misconception at every intermediate step, we ensure the reasoning steps can be achieved by students who hold that false belief, making the eventual contradiction at the last step more impactful in terms of the cognitive dissonance that it produces.

\begin{figure}[!t]
    \begin{newshaded}
    \small
    
    \noindent {\bf Your Task}
    
    You will be given a problem description, buggy code, a failed test case, and a student misconception. Your task is to write a reasoning trajectory: a sequence of rigorous, deductive reasoning steps that prove a statement contradicting the misconception.
    
    \vspace{0.5em}
    \noindent {\bf Core Principles}
    
    \begin{enumerate}[leftmargin=*,itemsep=2pt,topsep=3pt]
    \item {\bf Strictly deductive}: Each step must be a necessary logical consequence of previous steps, correct programming language knowledge, and observable facts.
    \item {\bf Consistent with misconception}: Do not assume programming knowledge that contradicts the student's false belief.
    \item {\bf Focus on disproving misconception}: End when reaching a statement that contradicts the misconception. Do not show the correct fix.
    \item {\bf Start from failed test}: Begin with observable facts from the failed test case and trace program state throughout execution.
    \item {\bf Sequential labeling}: Label steps as A.1, A.2, ..., A.n. Reference non-adjacent prior steps when used.
    \end{enumerate}
    
    \vspace{0.5em}
    \noindent {\bf Input Format}
    
\begin{verbatim}
<problem>[problem_description]</problem>
<bug_code>[buggy_code]</bug_code>
<failed_test>[failed_test]</failed_test>
<misconception>[misconception]</misconception>
\end{verbatim}
    
    \vspace{0.5em}
    \noindent {\bf Output Format}
    
\begin{verbatim}
Step A.1: [Observable fact(s) from failed test]
...
Step A.k: [Deduced fact(s) using previous steps]
...
Step A.n: [Statement contradicting misconception]
\end{verbatim}
    
    \end{newshaded}
    \caption{Prompt template for reasoning trajectory generation. The full template includes worked examples demonstrating code tracing and proof techniques such as loop invariants.}
    \label{fig:prompt-rt}
    \end{figure}

\subsection{Socratic Conversation Turns}
\label{sec:sc}

Building on the generated reasoning trajectories, we approach Socratic conversation generation as a sequential dialogue turn generation where each RT step anchors an instructor-student exchange. The instructor's utterances are intended to guide the student to make the inferences described at each RT step, instead of providing the inference step directly to the student. For example, if the RT step proves that \verb|range(1)| must have produced the value 0, the teacher should ask a question like \textcolor{blue}{\it "Where did the value 0 come from?"} rather than \textcolor{purple}{\it "Isn't it true then that {\tt range(1)} must have produced the value 0?"}.

Figure~\ref{fig:prompt-socratic} shows the prompt template for Socratic conversation generation. We employ a 1-shot prompting approach that includes a worked example showing the complete conversation associated with a reasoning trajectory. The prompt takes as input the complete reasoning trajectory along with the problem specification. The generated conversations follow a natural dialogue structure where the teacher begins by inquiring about the encountered issue, and subsequent turns systematically work through each RT step. Each teacher utterance corresponds to one RT step, aiming to elicit from the student the statement proven at that step. This one-to-one correspondence with the underlying reasoning trajectory ensures that the dialogue maintains logical coherence while preserving the pedagogical value of instructor-guided discovery.

\begin{figure}[!t]
    \begin{newshaded}
    \small
    
    \noindent {\bf Your Task}
    
    You will be given a Reasoning Trajectory (RT), which is a sequence of reasoning steps ending with a statement that disproves a student's misconception. Your task is to write a Socratic conversation between a Teacher and a Student that guides the student to articulate, at each turn, the statement proven at that RT step. The Teacher should not provide statements directly but ask questions that prompt the student to infer them independently.
    
    \vspace{0.5em}
    \noindent {\bf Guidelines}
    
    \begin{itemize}[leftmargin=*,itemsep=2pt,topsep=3pt]
    \item {\bf Natural conversation}: Teacher utterances should be direct, clear, and concise. Avoid phrases like ``That's an interesting point'' or ``Good question.''
    \item {\bf Socratic approach}: Ask open-ended questions that require reasoning. Do not state the inference and ask for confirmation.
    \item {\bf RT correspondence}: Each Teacher utterance prompts step A.X, and each Student response corresponds to A.X.
    \end{itemize}
    
    \vspace{0.5em}
    \noindent {\bf Formatting and Structure}
    
    \begin{itemize}[leftmargin=*,itemsep=2pt,topsep=3pt]
    \item Use {\tt Teacher:} and {\tt Student:} as speaker labels
    \item Conversation begins with Teacher inquiring about the issue
    \end{itemize}
    
    \vspace{0.5em}
    \noindent {\bf Input Format}
    
    \begin{verbatim}
<problem>[problem_description]</problem>
<buggy_code>[buggy_code]</buggy_code>
<failed_test>[failed_test]</failed_test>
<misconception>[misconception]</misconception>
<rt>[reasoning_trajectory]</rt>
\end{verbatim}
    
    \end{newshaded}
    \caption{Prompt template for Socratic conversation generation. The full template includes a worked example showing the correspondence between RT steps and dialogue turns.}
    \label{fig:prompt-socratic}
    \end{figure}

\section{Experimental Evaluation}
\label{sec:experiments}

We benchmark six state-of-the-art LLMs on their ability to generate valid reasoning trajectories and Socratic conversations: \texttt{GPT-5}, \texttt{GPT-5-mini}, \texttt{Claude Sonnet-4.5}, \texttt{Claude Haiku-4.5}, \texttt{Gemini 2.5-flash}, and \texttt{Gemini 2.5-pro}. The 6 LLMs are evaluated in 14 total configurations with varying levels of reasoning and different hyperparameters. All experiments leverage the API from the respective LLM providers.

\subsection{Model Configurations}

Our experiments evaluate three state-of-the-art LLMs via their respective APIs, each configured with model-specific parameters.

\noindent{\bf OpenAI GPT-5.} We employ the \texttt{gpt-5} and \texttt{gpt-5-mini} models using the Responses API with built-in reasoning capabilities at three effort levels: \textit{minimal}, \textit{low}, and \textit{medium}. We set max\_output\_tokens to 4000 and configure text verbosity to \textit{medium}.

\noindent{\bf Anthropic Claude.} We utilize \texttt{sonnet-4-5} and \texttt{haiku-4-5} with temperature 0.1 and max\_tokens 4000 for standard generation. When extended thinking is enabled, we increase temperature to 1.0, allocate an additional 2000 tokens for the thinking budget, and activate thinking mode with budget\_tokens set to 2000.

\noindent{\bf Google Gemini.} We use \texttt{gemini-2.5-flash} and \texttt{gemini-2.5-pro} with temperature 0.1 and max\_output\_tokens 4000 for baseline experiments. We set max\_output\_tokens to 6000 tokens and configure thinking\_config with include\_thoughts enabled and thinking\_budget set to 2000 in reasoning-enabled experiments.

\subsection{LLM-as-Judge Methodology}
\label{sec:llm-as-judge}

The sheer number of generated RT steps, over 22K as indicated in Table~\ref{tab:benchmark-stats}, prohibits comprehensive manual evaluation. Consequently, for both RT and Socratic conversation evaluation, we turn to using an LLM-as-judge approach, where:
\begin{enumerate}
    \item A suitably instructed LLM is first shown to be a reliable evaluator by manually verifying its decision on a small subset of examples (Section~\ref{sec:judge:evaluating}).
    \item The LLM is then deployed to automatically evaluate all generated trajectories and conversations (Section~\ref{sec:judge:using}).
\end{enumerate}
A priori, using the LLM-as-judge for LLM-based generations is sensible considering that, in general, {\it verification is much easier than generation}, e.g., determining whether a sequence of reasoning steps is sound is much easier than generating a sequence of reasoning steps that disproves a misconception.

\begin{table*}[t]
\centering
\begin{tabular}{lc|cc|cc}
\toprule
\textbf{Language Model} & \textbf{Reasoning} & \textbf{RT Steps} & \textbf{\% Valid RTs} & \textbf{\% Valid Convs} & \textbf{\% Grounded Turns} \\
\midrule
GPT-5 (minimal-effort) & \checkmark & 1,577 & 85.0\% & 94.3\% & 98.6\% \\
GPT-5 (low-effort) & \checkmark & 1,488 & 90.7\% & \textbf{98.7\%} & \textbf{99.4\%} \\
GPT-5 (medium-effort) & \checkmark & 1,271 & \textbf{91.1\%} & 94.8\% & 98.5\% \\
\midrule
GPT-5-mini (minimal-effort) & \checkmark & 1,826 & 68.3\% & 85.0\% & 96.9\% \\
GPT-5-mini (low-effort) & \checkmark & 1,453 & 59.5\% & 92.1\% & 98.0\% \\
GPT-5-mini (medium-effort) & \checkmark & 1,351 & 68.9\% & 95.9\% & 98.7\% \\
\midrule
Claude Sonnet-4.5 & \texttimes & 1,792 & 80.6\% & 89.0\% & 97.4\% \\
Claude Sonnet-4.5 & \checkmark & 1,776 & 87.2\% & 92.5\% & 97.9\% \\
\midrule
Claude Haiku-4.5 & \texttimes & 1,962 & 62.6\% & 68.3\% & 93.2\% \\
Claude Haiku-4.5 & \checkmark & 1,738 & 78.9\% & 81.5\% & 95.7\% \\
\midrule
Gemini 2.5-flash & \texttimes & 1,379 & 83.5\% & 86.1\% & 97.4\% \\
Gemini 2.5-flash & \checkmark & 1,826 & 82.7\% & 85.8\% & 96.7\% \\
\midrule
Gemini 2.5-pro & \texttimes & 1,439 & 77.2\% & 78.3\% & 94.9\% \\
Gemini 2.5-pro & \checkmark & 1,628 & 85.3\% & 78.7\% & 95.5\% \\
\bottomrule
\end{tabular}
\caption{Performance of language models on reasoning trajectory generation and Socratic conversation generation. RT Steps shows total steps across all 227 samples. Valid Convs measures whether all teacher turns in a conversation are grounded in the RT; Grounded Turns measures the percentage of individual teacher turns that are properly grounded.}
\label{tab:rt-socratic-results}
\end{table*}

\subsubsection{Using the LLM-as-Judge}
\label{sec:judge:using}

To evaluate the RT generation step, we employ the LLM-as-judge approach with structured criteria across four major categories: logical soundness, step construction, precision, and focus, where a correct RT must satisfy all criteria. We then compute the percentage of correct RTs for each model. Figure~\ref{fig:prompt-rt-eval} shows the evaluation criteria.

To evaluate the quality of generated Socratic turns, we employ an LLM-as-judge approach as well, with two key criteria: whether the teacher utterance elicits the correct inference from the corresponding RT step, and whether it avoids stating that inference directly. For a teacher Socratic utterances to be deemed correct, it must satisfy both criteria. We then compute the percentage of valid Socratic turns for each model. Lastly, we compute the percentage of valid Socratic conversations for each model, where a valid conversation must have all valid teacher utterances grounded in the corresponding RT step. Figure~\ref{fig:prompt-socratic-eval} shows the evaluation criteria for Socratic turns.

\subsubsection{Evaluating the LLM-as-Judge}
\label{sec:judge:evaluating}

To evaluate the reliability of the LLM-as-judge, we conducted a manual evaluation of the LLM-as-judge output on a subset of 30 RT samples, 10 from each of three models: Claude Sonnet-4.5 with reasoning, Gemini 2.5-pro with reasoning, and GPT-5 with medium reasoning effort. For each sample, we generated reasoning trajectories and Socratic conversations using the model configurations specified above. We then evaluated these outputs using Claude Sonnet-4.5 with extended thinking as the LLM judge (temperature 1.0, max\_tokens 8000), applying the evaluation criteria shown in Figures~\ref{fig:prompt-rt-eval} and~\ref{fig:prompt-socratic-eval}. An established programmer with knowledge of theorem proving then independently evaluated the same 30 samples for RT validity. Correspondingly, we observed a 76.7\% agreement between the LLM judge and the human expert on reasoning trace evaluation. However, this is a very conservative estimate, as the expert penalized the LLM-judge when it missed technical inaccuracies in terminology, such as using the term "conditional expression" to mean a boolean condition when it actually refers to Python's ternary operator \texttt{x if C else y}. Such cases do not significantly impact the soundness of the RT, and in practice can still be useful for articulating valid Socratic conversations.

We also use a subset of 88 teacher utterances to manually evaluate the Socratic turn quality using the same criteria as the LLM-as-judge. Correspondingly, we observed a 96.6\% agreement on Socratic turn evaluation. 
In Socratic turn validation, the judge demonstrates strong reliability on clear-cut cases and consistently detects and penalizes teacher utterances consisting of rhetorical questions seeking confirmation. The judge occasionally makes evaluation mistakes whereby it penalizes useful conversational framing, e.g., {\it "Let's trace through the code"}, by motivating that it is not relevant to eliciting the target reasoning step from the student.

\begin{figure}[!h]
    \begin{newshaded}
    \small
    
    \noindent {\bf Your Task}
    
    Evaluate whether a reasoning trajectory (RT) serves as a rigorous, logical proof by counterexample that contradicts a student misconception. An RT is {\bf VALID} only if it passes all criteria in all three categories below.
    
    \vspace{0.5em}
    \noindent {\bf Category 1: Logical Soundness}
    
    \begin{itemize}[leftmargin=*,itemsep=0pt,topsep=3pt]
    \item {\bf Valid Starting Point}: Begins with verifiable fact from failed test.
    \item {\bf Deductively Valid}: Each step follows necessarily from prior steps and Python semantics. No abduction or logical leaps. Steps do not assume programming knowledge that directly contradicts the misconception.
    \item {\bf Sound Contradiction}: Establishes facts incompatible with misconception.
    \item {\bf Complete Causal Chain}: Unbroken chain from observation to contradiction.
    \item {\bf Execution Tracing}: Trace program execution to deduce concrete facts.
    \end{itemize}
    
    \vspace{0.5em}
    \noindent {\bf Category 2: Step Construction \& Precision}
    
    \begin{itemize}[leftmargin=*,itemsep=0pt,topsep=3pt]
    \item {\bf Clear Boundaries}: Each step is a distinct logical unit
    \item {\bf Precision}: Uses specific line numbers, variable names, values
    \item {\bf Proper Citation}: Non-adjacent dependencies explicitly cited
    \item {\bf Technical Accuracy}: All claims about Python constructs are correct
    \end{itemize}
    
    \vspace{0.5em}
    \noindent {\bf Category 3: Formatting \& Focus}
    
    \begin{itemize}[leftmargin=*,itemsep=0pt,topsep=3pt]
    \item {\bf Sequential Labeling}: All steps labeled sequentially (A.1, A.2, ...)
    \item {\bf Focus on Misconception}: Exclusively focused on disproving target misconception
    \end{itemize}
    
    \vspace{0.5em}
    \noindent {\bf Output Format}
    
    \begin{verbatim}
    {
      "valid": true/false,
      "categories": {
        "logical_soundness": true/false,
        "step_construction_and_precision": true/false,
        "formatting_and_focus": true/false
      },
      "comments": "[Evaluation rationale]",
      "feedback": "[Actionable suggestions or NONE]"
    }
    \end{verbatim}
    \end{newshaded}
    \caption{Prompt template for LLM-as-judge evaluation of RTs. An RT is considered valid only if all three categories pass.}
    \label{fig:prompt-rt-eval}
    \end{figure}

\begin{figure}[!h]
    \begin{newshaded}
    \small
    
    \noindent {\bf Your Task}
    
    Evaluate whether a Teacher utterance in a Socratic conversation effectively guides a student to articulate the inference from a specific RT step. A teacher utterance is {\bf VALID} only if it satisfies both criteria below.
    
    \vspace{0.5em}
    \noindent {\bf Criterion 1: Prompts the Correct Inference}
    
    The teacher's question must guide the student to articulate the key inference from the specific RT step it claims to prompt. The student's response should contain the statement proven in that step, and only that step. Questions may state facts from previous steps but must prompt the new inference at the target step.
    
    \vspace{0.5em}
    \noindent {\bf Criterion 2: Does Not State the Inference Directly}
    
    The teacher must ask a question requiring reasoning. The teacher should not provide the answer or state the conclusion. Questions can be general (``What's the issue?'') or specific, as long as they require the student to think and derive the answer rather than merely confirm a stated fact.
    
    \vspace{0.5em}
    \noindent {\bf Evaluation Process}
    
    \begin{enumerate}[leftmargin=*,itemsep=2pt,topsep=3pt]
    \item Read RT step A.X to understand the target inference
    \item Read RT steps A.1 through A.X-1 for established facts
    \item Read the teacher utterance and student response
    \item Evaluate against both criteria
    \item Valid only if both criteria pass
    \end{enumerate}
    
    \vspace{0.5em}
    \noindent {\bf Output Format}
    
    \begin{verbatim}
    {
      "valid": true/false,
      "criteria_scores": {
        "prompts_correct_inference": true/false,
        "does_not_state_inference": true/false
      },
      "comments": "[Evaluation explanation]",
      "feedback": "[Suggestions or NONE]"
    }
    \end{verbatim}
    
    \end{newshaded}
    \caption{Prompt template for LLM-as-judge evaluation of Socratic utterances. An utterance is considered valid only if it prompts the student to make the correct reasoning step without stating the inference directly.}
    \label{fig:prompt-socratic-eval}
    \end{figure}

\subsection{Quantitative Results}

The results from all 14 LLM configurations are summarized in Table~\ref{tab:rt-socratic-results} and reveal several key findings. First, reasoning trajectory quality varies considerably, with GPT-5 achieving the highest validity rates between $85-91\%$, while generating relatively concise trajectories. Notably, extended reasoning capabilities do not uniformly improve performance. Claude models benefit substantially from reasoning mode: Claude Sonnet-4.5 with $+6.6\%$ in RT validity and Claude Haiku-4.5 with $+16.3\%$ in RT validity. Similarly, GPT-5 models perform better with increased reasoning effort. In contrast, the results from Gemini models are mixed, with 2.5-flash performing slightly worse when reasoning is enabled, with $-0.8\%$ in RT validity, while generating more reasoning steps.

We also observe a slight inverse relationship between trajectory length and validity: GPT-5 medium-effort produces the fewest total steps (1,271) and achieves the highest RT validity (91.1\%), while Claude Haiku-4.5 without reasoning generates the most steps (1,962) but has the lowest validity (62.6\%). This is somewhat to be expected, given that the more reasoning steps are contained in an RT, the more chances for one of them to be invalid, which then, according to our evaluation methodology, invalidates the entire RT.

Generally, once a reasoning trajectory is generated, the Socratic conversation generation process is relatively straightforward and consistent across different models. Most LLMs are able to generate valid Socratic utterances grounded in the input reasoning trajectory, and they consistently do so throughout an entire conversation.

\subsection{Qualitative Analysis}
\label{sec:analysis:}

To complement the quantitative evaluation, we conducted qualitative analysis of RT generation on 30 manually evaluated samples from all of the LLM configurations. We identified key patterns characterizing successful and unsuccessful reasoning trajectories.

\subsubsection{Success Pattern: Exhausting Alternative Possibilities}

Effective reasoning trajectories enumerate all possible scenarios and systematically eliminate those contradicting observed behavior. Consider an example where a student has written:

\begin{lstlisting}[language=Python]
def calculate_average(x, y):
    return x + y / 2
\end{lstlisting}

The student incorrectly believes that \texttt{+} has higher precedence than \texttt{/}. When called as \texttt{calculate\_average(1, 3)}, the function returns 2.5 instead of the expected 2.0. The reasoning trajectory establishes:

{\bf Step A.1:} The failed test states that \texttt{calculate\_average(1, 3)} returns 2.5. So with \texttt{x = 1} and \texttt{y = 3}, the expression on line 2, \texttt{x + y / 2}, evaluates to 2.5.

{\bf Step A.2:} There are no parentheses in line 2, so the only two possible groupings of \texttt{x + y / 2} are: (1) \texttt{(x + y) / 2}, or (2) \texttt{x + (y / 2)}.

{\bf Step A.3:} Compute \texttt{(x + y) / 2} with \texttt{x = 1} and \texttt{y = 3}: \texttt{(1 + 3) / 2 = 4 / 2 = 2.0}.

{\bf Step A.4:} If \texttt{+} had higher precedence than \texttt{/}, then line 2 would be evaluated as \texttt{(x + y) / 2}, which we computed to be 2.0 (A.3). But the actual result is 2.5 (A.1). Therefore, \texttt{+} is not evaluated before \texttt{/} in this expression.

This approach demonstrates systematic elimination: the RT enumerates all possible interpretations (Step A.2), computes the result under each interpretation (Step A.3), and rules out the interpretation matching the misconception by contrasting it with observed behavior (Step A.4). 

\subsubsection{Success Pattern: Concrete Execution Tracing}
Successful reasoning trajectories ground abstract reasoning in concrete test values, tracing program execution with specific inputs throughout the logical chain. In the same \texttt{calculate\_average} example above, the RT uses the specific values \texttt{x = 1} and \texttt{y = 3} from the failed test consistently across all steps. Step A.1 establishes these concrete values from the failed test. Step A.3 computes the concrete result \texttt{(1 + 3) / 2 = 4 / 2 = 2.0} for the first grouping. Step A.5 (not shown above) verifies the alternative grouping \texttt{1 + (3/2) = 2.5} matches the observed output. This concrete tracing ensures every deductive step is verifiable against observable program behavior rather than relying on abstract reasoning about Python semantics. The specificity eliminates ambiguity and makes each logical inference checkable.

\subsubsection{Success Pattern: Clear Contradiction}

Effective reasoning trajectories explicitly structure the contradiction between the misconception and observed behavior. Step A.4 in the example above demonstrates this structure: it first states the implication of the misconception (``If \texttt{+} had higher precedence than \texttt{/}, then line 2 would be evaluated as \texttt{(x + y) / 2}"), then computes the result under that assumption (``which we computed to be 2.0"), and finally contrasts this with actual observed behavior (``But the actual result is 2.5"). The explicit ``If...then...But" structure makes the logical contradiction transparent. This pattern ensures the reasoning trajectory achieves its primary purpose: proving the misconception leads to predictions incompatible with observed program behavior.

\subsubsection{Failure Pattern: Using Knowledge that Contradicts the Misconception}

The most pedagogically damaging failure pattern occurs when reasoning trajectories rely on programming knowledge that directly contradicts the target misconception. Consider an example where a student has written:

\begin{lstlisting}[language=Python]
def top_k(lst, k):
    result = []
    for i in range(k):
        result.append(max(lst))
        lst.pop(max(lst))  # Line 5
    return result
\end{lstlisting}
The student incorrectly believes that the \texttt{.pop()} method takes a value to be deleted from the list. When called as \texttt{top\_k([1, 2, 3, 4, 5], 1)}, the function raises an \texttt{IndexError} on line 5. The reasoning trajectory contains:

{\bf Step A.5:} In Python, \texttt{list.pop(i)} removes and returns the item at index \texttt{i}; if \texttt{i} is outside the valid index range for the list, it raises \texttt{IndexError}. Therefore, the observed \texttt{IndexError} on \texttt{lst.pop(5)} means the argument 5 was used as an index, not as a value.

This step explicitly states how \texttt{list.pop()} actually works, directly contradicting the student's belief. A student holding the misconception would thus be directly provided the cause of the bug, which defeats the purpose of a Socratic approach. The RT evaluation criterion states that reasoning steps must not assume programming knowledge that directly contradicts the misconception. The correct approach would prove that interpreting the argument as a value leads to a contradiction, without stating how \texttt{pop()} actually works.

\subsubsection{Failure Pattern: Abductive Reasoning and Logical Leaps}

Reasoning trajectories sometimes employ abductive reasoning (inference to the best plausible explanation) rather than deductive proof, leaving logical gaps. Consider the student code sample below:

\begin{lstlisting}[language=Python]
def count_words(sentence):
    words = 0
    space_mode = True
    for i in range(1, len(sentence)):  # Line 4
        if sentence[i] == ' ':
            if not space_mode:
                words += 1
            space_mode = True
        else:
            space_mode = False
    if not space_mode:
        words += 1
    return words
\end{lstlisting}

The student incorrectly believes string indexing starts at 1. When called as \texttt{count\_words("I love Python")}, the function returns 2 instead of 3. After establishing that the loop executes from \texttt{i = 1} to \texttt{i = 12} and that \texttt{words} is incremented exactly once during the loop, the reasoning trajectory contains:

{\bf Step A.9:} At the start of the loop when \texttt{i = 1}, \texttt{space\_mode} is \texttt{True} (initialized on line 3). If \texttt{sentence[1]} is a space, line 6's condition \texttt{not space\_mode} would be \texttt{False}, so line 7 would not execute. For the algorithm to eventually count only 2 words while ``I love Python" has 3 words, and for line 7 to execute exactly once, \texttt{sentence[1]} must be a space.

This reasoning shows that if \texttt{sentence[1]} a space, the code produces \texttt{words = 2}, and uses that to conclude that \texttt{sentence[1]} must be space. But this does not show that \texttt{words = 2} necessarily implies that \texttt{sentence[1]} is a space.

Another example demonstrates this pattern more concisely. In the earlier \texttt{top\_k} example, a different reasoning trajectory contains:

{\bf Step A.7:} Since calling \texttt{lst.pop(5)} raises an \texttt{IndexError}, and the number 5 is an invalid index for \texttt{lst}, the \texttt{.pop()} method must be interpreting the argument 5 as an index, not as a value.

The phrase ``must be interpreting" reveals abduction. The step infers the most likely explanation but does not prove it deductively. It is theoretically possible for \texttt{pop()} to use the argument in a different way while still raising \texttt{IndexError}. A valid approach would need to prove that if 5 were interpreted as a value to remove, no error would occur (since 5 exists in the list), establishing contradiction.

\subsubsection{Socratic Utterance Quality}

We conducted qualitative analysis of Socratic utterance generation on 88 teacher utterances from 15 reasoning trajectories, 5 from each of three models: Claude Sonnet-4.5 with reasoning, Gemini 2.5-pro with reasoning, and GPT-5 with medium reasoning effort. We identified four key success patterns:

(1) {\bf Accurate RT step alignment}: We observed no cases where a teacher question prompted content from a completely different RT step than requested, demonstrating that LLMs reliably understand the step-by-step structure of reasoning trajectories.

(2) {\bf Step reference integration}: Effective questions successfully synthesize facts established in prior RT steps while prompting a new inference step through a single, coherent question;

(3) {\bf Implicit elicitation}: Questions prime students to provide complete logical steps beyond what is explicitly requested. For instance, asking \textcolor{blue}{\it "What expression is evaluated on line 5?"} implicitly elicits both the abstract expression (\texttt{sen.replace(i, ``x")}) and its concrete evaluation with values substituted (\texttt{`a'.replace(`a', `x')}), without requiring separate prompts for each component;

(4) {\bf Generic opening questions}: All conversations begin with broad, open-ended questions that effectively elicit students' initial observations about failed tests, without leading them.

The most common failure pattern observed is where teacher utterances state conclusions directly rather than prompting students to derive them. For example, stating {\it "Your assumption that addition comes first leads to a final result of 2.0"} before asking {\it "How does that compare to what the program actually calculated?"} provides the conclusion, reducing the students' task to mere confirmation rather than more substantive reasoning.

\section{Streamlit Web Interface}
\label{sec:web-interface}

To support practical use of our approach, we developed an interactive web application using Streamlit that implements an end-to-end pipeline: from a problem specification and buggy student code to RTs and complete Socratic debugging conversations. The application runs locally and securely loads API credentials from environment variables.




\begin{figure}[!b]
    \centering
\includegraphics[width=\linewidth]{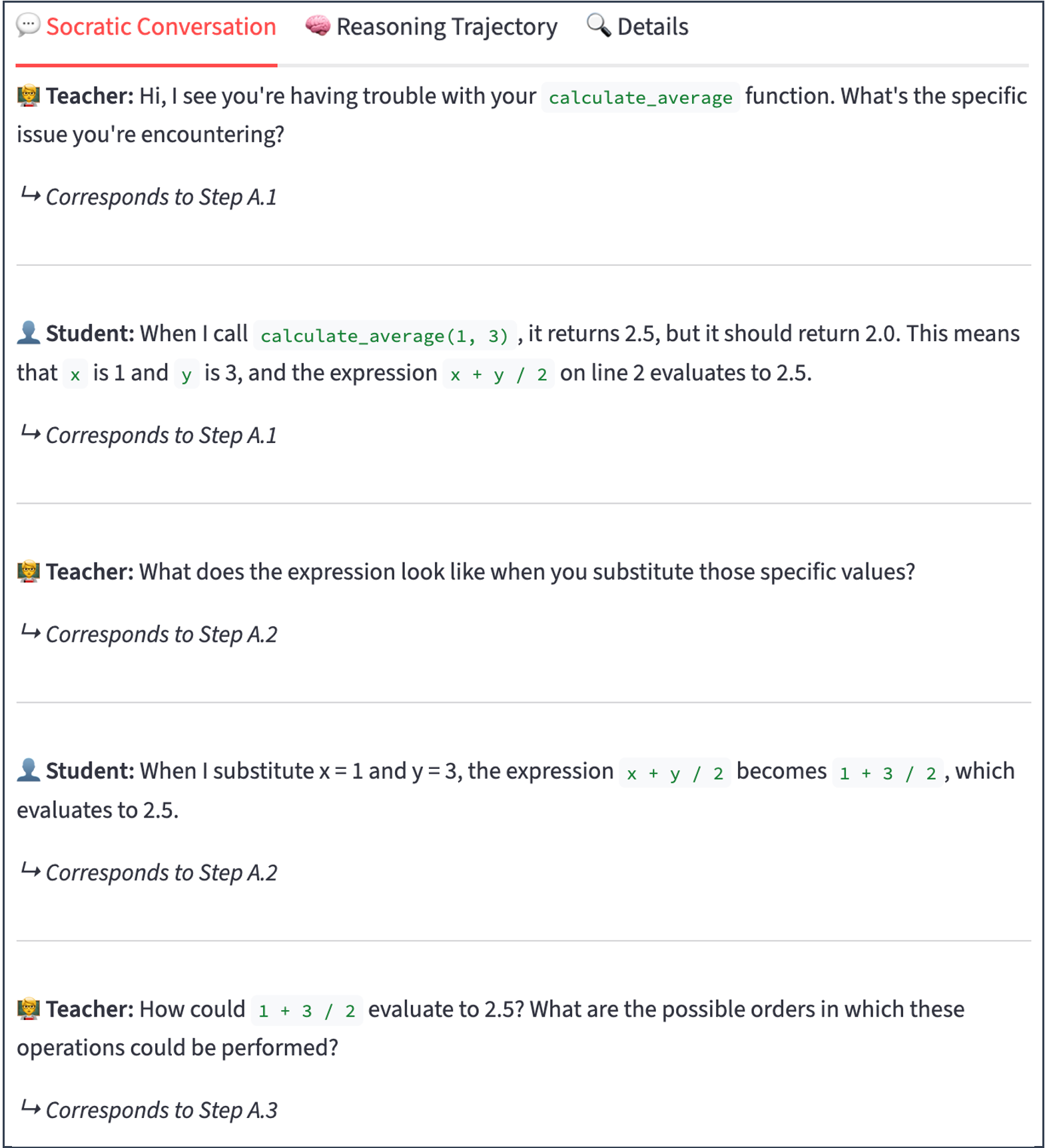}
    \caption{Snapshot from the web interface, showing the first part of a Socratic conversation between a student and a teacher based on an already generated reasoning trajectory.}
    \label{fig:interface-dialogue}
\end{figure}

The workflow begins when users input a problem description along with student code that fails one or more test cases, a description of the failed test case, and a misconception description. The application generates a reasoning trajectory that maps a path from the student's flawed mental model to a contradictory statement about program behavior. This trajectory subsequently guides the generation of a Socratic conversation demonstrating how an instructor might guide the student toward self-discovery of their error.

Instructors can select from multiple state-of-the-art LLMs (Claude, GPT, Gemini), with the interface automatically applying the same prompt templates and model configurations used in our experimental evaluation. Reasoning capabilities are enabled by default for all compatible models. The interface presents results in a structured format featuring: the complete reasoning trajectory with intermediate steps, an example Socratic conversation, and expandable reasoning traces showing the model's internal deliberation process. An instructor can use directly the generated Socratic conversation, or they can articulate their own utterances for each step in the reasoning trajectory.

For classroom applications, instructors can use this tool to rapidly prepare targeted interventions for individual students. When a student presents buggy code, the instructor can input it into the system and within seconds receive a principled Socratic questioning strategy tailored to a specific misconception. This enables more effective one-on-one debugging sessions, wherein the instructor follows a structured reasoning plan rather than ad-hoc questioning.

\section{Limitations and Threats to Validity}

Due to the substantial cognitive effort and time involved, the failure cases revealed by the qualitative analysis in Section~\ref{sec:analysis:} were extracted from a manual analysis of only 30 LLM-generated samples. Similarly, the LLM-as-judge solution for evaluation in Section~\ref{sec:llm-as-judge}, albeit justified by the very large number of generated samples, can be prone to some false positive and false negatives, especially when judging the soundness of reasoning trajectories. We hope these limitations to be mitigated by the open source nature of the project and its web interface, which will allow anyone to use and stress-test the approach. Furthermore, the evaluation assumes a static setting, where the student successfully follows the instructor along the reasoning trajectory. Real conversations, however, can lead to situations where the student struggles to make the next inference step, requiring the instructor to change strategy mid-conversation. We leave the modeling of this dynamic setting for future work.

\section{Related Work}
\label{sec:related}

Scaffolding enables learners to achieve goals through guided efforts~\cite{wood1976role}, whereas Socratic Questioning represents a conversational form of scaffolding~\cite{wood1976role,quintana2018scaffolding,vygotsky2012thought}. Wood~\citeyearpar{wood1994patterns} identified two key questioning types: funneling, which guides learners toward solutions through sequential questions, and focusing, which directs attention to important problem aspects and encourages reflection~\cite{wood1994patterns,national_council_of_teachers_of_mathematics_principles_2014,alic2022computationally}. While students can complete programming exercises yet struggle to explain their code~\cite{lehtinen2021students}, Tamang et al.~\citeyearpar{tamang2021_socratic} demonstrated that Socratic methods effectively improve code comprehension. 

Prior AI work in programming education includes intelligent tutoring systems (ITS) and learning support systems that provide automated feedback, generate exercises, and create code explanations~\cite{sami_2022_codex_edu}. Most ITS models use pre-LLM methods like action-rules and Bayesian networks~\cite{crow2018intelligent, mousavinasab2021intelligent, costello_2012_adaptive, butz2006web}. Recent work has shown that computer-based scaffolding techniques have a moderate impact on STEM learning~\cite{kim2018effectiveness}. Some approaches use fine-tune LMs to automatically generate Socratic questions for math problems~\cite{shridhar-etal-2022-automatic, macina-etal-2023-mathdial, macina2025mathtutorbench}. Furthermore, several open source LLMs have been fine-tuned on synthetic tutoring conversations in mathematics~\cite{socraticLM2025} and on real tutoring conversations in multiple subjects~\cite{perczel2025teachlm}.

Automated hint generation systems aim to assist programming students through instant feedback using techniques like extracting common bugs~\cite{lee2018vida}, analyzing peer data patterns~\cite{iii2014generating, lazar2017automatic}, and generating custom solution paths~\cite{rivers2017data, mcbroom2021survey}. AI tutoring for formal proving in mathematics, such as the LeanTutor~\cite{patel2025leantutor}, rely on generating three types of hints: an identification of the error, a single guiding question, an explicit suggestion for the next step; thus, they do not engage in a complete Socratic conversation. Lu and Krishnamurthi~\citeyearpar{lu_2024_krishnamurthi} present an approach to identifying and correcting student misconceptions about programming language behavior through "misinterpreters", pre-programmed interpreters that can deterministically detect misconceptions about programming language semantics. Their SMoL Tutor uses refutation texts to explicitly address these misconceptions during MCQ quizzes.


Our approach focuses on the diagnosis and correction of misconceptions in buggy code through complete Socratic dialogue. Unlike prior work, we plan Socratic conversations such that they engage the student in a particular type of reasoning about the buggy code behavior, where they are guided towards inferring a correct statement about the actual code execution that conflicts with their misconception. As argued in Section~\ref{sec:introduction}, reaching this moment of cognitive dissonance is expected to trigger an enduring belief update that fixes the misconception.


\section{Conclusion and Future Work}
\label{sec:conclusion}

We introduced a novel formulation of Socratic debugging, where the teacher utterances aim to follow a reasoning trajectory that starts from a failed test case, and upon a sequence of inference steps reaches a correct statement about the program that is in contradiction with the student misconception that caused the bug. Upon reaching this statement, the student is expected to experience a strong cognitive dissonance, which then entails an enduring belief update. To support development and evaluation, we created a dataset of 227 problems paired with buggy solutions and the corresponding bug-causing misconception. A large scale LLM-as-judge evaluation of over 22K generated reasoning trajectory steps and their associated Socratic utterances shows that large language and reasoning models can achieve up to 91\% trajectory validity and 98.7\% conversation validity. Overall, through carefully orchestrated moments of cognitive dissonance, the proposed automated Socratic guidance approach can be of significant benefit to instructors seeking to help students durably fix their programming misconceptions.

In future work, we plan to develop approaches for simplifying the input to Socratic debugging in order to lessen the cognitive demands on the student by focusing on the code behavior that contains the misconception while staying as close to the original buggy code as possible.

\bibliographystyle{ACM-Reference-Format}
\bibliography{references}



\end{document}